\documentclass{article} 
\usepackage{iclr2024_conference,times}

\usepackage{amsmath,amsfonts,bm}

\def\eqref#1{equation~\ref{#1}}

\def\1{\bm{1}}

\DeclareMathAlphabet{\mathsfit}{\encodingdefault}{\sfdefault}{m}{sl}
\SetMathAlphabet{\mathsfit}{bold}{\encodingdefault}{\sfdefault}{bx}{n}

\usepackage{hyperref}
\usepackage{url}
\usepackage{enumitem}
\usepackage{graphicx}

\usepackage{amssymb}
\usepackage{makecell}
\usepackage{amsmath}
\usepackage{xspace}
\usepackage{booktabs}
\usepackage{multirow, makecell} 
\usepackage{adjustbox}
\usepackage{wrapfig,lipsum,booktabs}
\usepackage[breakable]{tcolorbox}

\usepackage{tikz}
\usepackage{pgfplots}
\usepgfplotslibrary{statistics}
\usetikzlibrary{plotmarks}
\usepackage{wrapfig}

\usepackage{pifont}
\definecolor{lightergray}{RGB}{230,230,230}
\definecolor{DarkGreen}{RGB}{30,130,30}

\newcommand\ours{\textsc{GraphText}\xspace}

\definecolor{myblue}{RGB}{0, 133, 186}
\definecolor{mygreen}{RGB}{72, 194, 0}
\definecolor{myred}{RGB}{177, 0, 28}

\title{GraphText: Graph Reasoning in Text Space}

\author{Jianan Zhao$^{1,2}$,\, Le Zhuo$^{3}$,\, Yikang Shen$^{4}$,\, Meng Qu$^{1,2}$,\, Kai Liu$^{5}$\\ \textbf{\text{Michael Bronstein}$^{6}$,\, Zhaocheng Zhu$^{1,2}$,\, Jian Tang$^{1,7,8}$} \\
    $^1$Mila - Qu\'ebec AI Institute,\,$^2$Universit\'e de Montr\'eal,\, $^3$Beihang University,\\
    $^4$MIT-IBM Watson AI Lab,\,$^5$Division of gRED Computational Science, Genentech Inc.,\\
    $^6$University of Oxford, \,$^7$HEC Montr\'eal,\,
    $^8$Canadian Institute for Advanced Research (CIFAR) \\
}

\iclrfinalcopy 
\begin{document}
\vspace{-15pt}
\maketitle
\vspace{-5pt}
\begin{abstract}
Large Language Models (LLMs) have gained the ability to assimilate human knowledge and facilitate natural language interactions with both humans and other LLMs. However, despite their impressive achievements, LLMs have not made significant advancements in the realm of graph machine learning. This limitation arises because graphs encapsulate distinct relational data, making it challenging to transform them into natural language that LLMs understand. In this paper, we bridge this gap with a novel framework, \ours, that translates graphs to natural language. \ours derives a graph-syntax tree for each graph that encapsulates both the node attributes and inter-node relationships. Traversal of the tree yields a graph text sequence, which is then processed by an LLM to treat graph tasks as text generation tasks. Notably, \ours offers multiple advantages. It introduces {\em training-free graph reasoning}: even without training on graph data, \ours with ChatGPT can achieve on par with, or even surpassing, the performance of supervised-trained graph neural networks through in-context learning (ICL). Furthermore, \ours paves the way for  {\em interactive graph reasoning}, allowing both humans and LLMs to communicate with the model seamlessly using natural language. These capabilities underscore the vast, yet-to-be-explored potential of LLMs in the domain of graph machine learning.
\end{abstract}
\vspace{-5pt}

\section{Introduction}

Language stands as a cornerstone of human civilization, acting as the primary medium for knowledge encoding, reasoning, and communication. Large language models (LLMs), pre-trained on extensive text corpora, have showcased remarkable reasoning skills~\citep{GPT3,GPT4}. These LLMs can communicate via natural language both internally~\citep{CoT} and externally with humans or other LLMs~\citep{CAMEL}, demonstrating exceptional skills such as multi-step reasoning~\citep{ToT}, decision-making~\citep{ReAct, CodeAsPolicies}, tool use~\citep{Toolformer}, and multi-agent collaboration~\citep{GenerativeAgents,MetaGPT}.

\paragraph{Motivation.} Despite the remarkable success of LLMs in handling natural languages, their application to other data modalities presents unique challenges, primarily because these data often lack straightforward transformation into sequential text. These challenges are especially severe when dealing with graph-structured data, as different graphs define structure and features in distinct ways. Therefore, existing efforts within the graph machine learning field commonly require the training of specific graph neural networks (GNNs) tailored to individual graphs~\citep{GCN, GAT, GIN}. Often, models trained on one graph cannot generalize to the unseen structure and feature representations of other graphs. Moreover, the gap between graphs and human languages hinders the application of natural language reasoning to facilitate graph reasoning.

\begin{figure*}[t]
    \centering
    \includegraphics[width=0.98\linewidth]{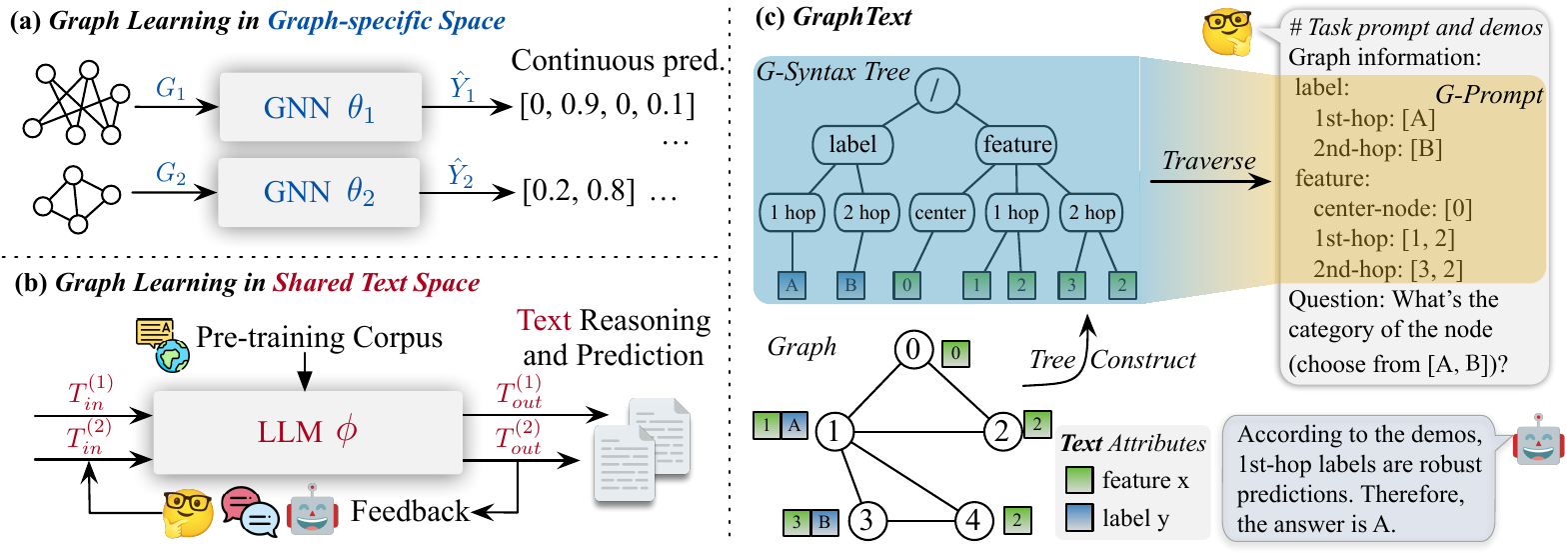}
    \caption{Comparison between \textbf{(a)} the GNN framework and \textbf{(b)} the proposed \ours framework. For different graphs $G_1$ and $G_2$, different GNNs $\theta_1$ $\theta_2$ are trained to make a graph-specific output prediction in continuous form. In contrast, \ours encodes the graph information to text sequences $T_{in}^{(1)}$ and $T_{in}^{(2)}$, and generates text reasoning and prediction $T_{out}^{(1)}$ and $T_{out}^{(2)}$ with a graph-shared LLM $\phi$. \ours leverages a pre-trained LLM to perform training-free graph reasoning and enables human and AI interaction for graph reasoning in natural language. \textbf{(c)} An example of the \ours framework that classifies node 0: Given a graph, \ours constructs a graph-syntax tree that contains both node attributes (e.g. feature and label) and relationships (e.g. center-node, 1st-hop, and 2nd-hop). Then, \ours traverses the graph-syntax tree to obtain a sequential text, i.e. graph prompt, and let LLM perform graph reasoning in text space.}
    \label{fig: GraphText}
\end{figure*}

In light of these limitations, a question arises: \textit{can we derive a language for graph in natural language?} In this paper, we give an affirmative answer by proposing to use \textit{tree} as an intermediary, elegantly bridging structured data and one-dimensional sequential language. Essentially, \textit{a tree exhibits a hierarchical structure, and traversing it yields a one-dimensional sequence.} On top of that, as shown in Figure \ref{fig: GraphText} (c), we propose a novel framework \ours, which takes graph data to build a graph-syntax tree. Traversing it results in a graph prompt expressed in natural language, allowing an LLM to approach graph reasoning as a text-generation task.



\begin{figure*}[t]
    \centering
    \includegraphics[width=0.95\linewidth]{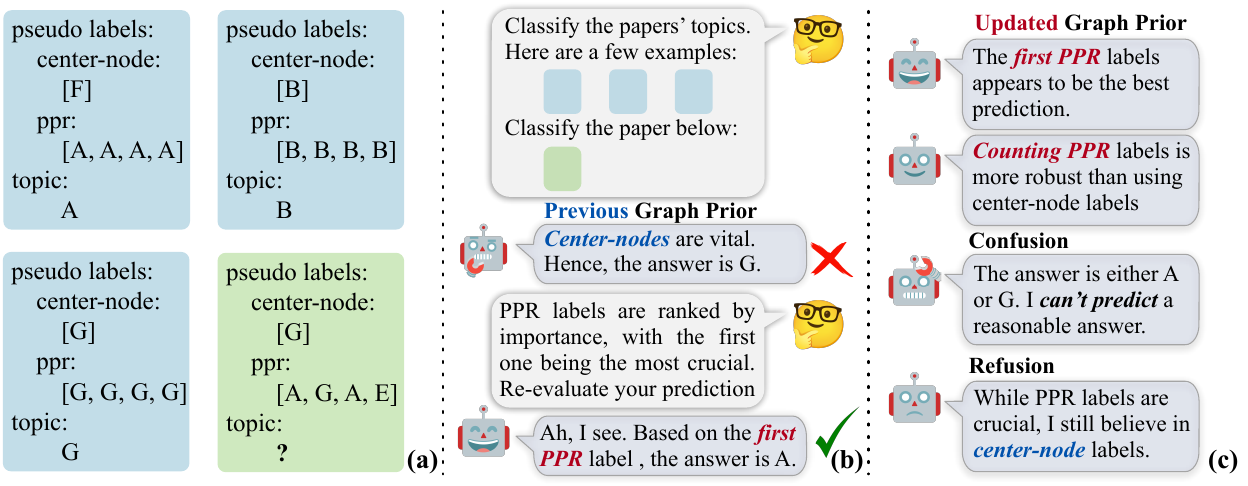}
    \caption{\textbf{(a)} Few-shot demonstrations \textcolor{myblue}{(blue)} and the target node \#2188 \textcolor{mygreen}{(green)} to predict on Cora. \textbf{(b)} An illustration of how human interaction changes the graph reasoning of an LLM, where the LLM previously has the prior that the center-node is vital. \textbf{(c)} Behaviors of LLMs after given demonstrations/human interaction: \textit{update graph prior} to bias more on PPR (personalized pagerank); leads to \textit{confusion} or \textit{refusion}. Details are discussed in Section \ref{sec: exp interactive}.
 }
    \label{fig: InteractWithHumanExample}
\end{figure*}

\vspace{-5pt}
\paragraph{Main contributions.} First, \ours serves as a flexible and general framework for graph reasoning. It can incorporate common inductive bias of GNNs, such as feature propagation and feature similarity-based propagation, simply by constructing different graph-syntax trees. It also serves as a general framework for graph reasoning for both in-context learning and instruction tuning, on both general graphs and text-attributed graphs.
%
Second, we show that \ours enables the possibility of \textbf{training-free graph reasoning}. The training-free property enables us to deploy \ours not only with open-source LLMs, but also with powerful closed-source LLMs. Remarkably, even without training on graph data, \ours with ChatGPT can deliver performance on par with, or even surpassing, supervised graph neural networks through in-context learning. This highlights the vast potential of foundation models in the realm of graph machine learning.
Third, \ours fosters \textbf{interactive graph reasoning}: With its capacity to generate and {\bf explain} predictions in natural language, humans can directly engage with \ours. As shown in Figure~\ref{fig: InteractWithHumanExample} (b), through interactions with humans and other LLMs, \ours refines its graph reasoning capabilities.

\section{Methodology}
\label{sec: graph language}

In this section, we introduce \ours to perform graph reasoning in text space. 
Out of the three fundamental problems of graph ML (graph classification, node classification, and link prediction), we 
take node classification as an example to introduce our idea. We however note that our discussion applies to other graph tasks.

\subsection{The \ours Framework}
\label{sec: model framework}
Let us be given an attributed graph $G=(V, E, \boldsymbol{X})$ with nodes $V$ and edges $E$, whose structure is represented as the $|V|\times |V|$ adjacency matrix $\boldsymbol{A}$
and node attributes as the $|V|\times d$ feature matrix $\boldsymbol{X}$. 
Given a subset $L \subset V$ of labeled nodes with labels $Y_L$, 
the goal of node classification is to predict the labels $Y_U$ of the unlabeled nodes $U = V\setminus L$. 
Graph Neural Networks (GNNs) are the standard architecture for such problems. 
As shown in Figure \ref{fig: GraphText} (a), a GNN directly learns a parametric map 
\begin{equation}
  \hat{y}_i =f_{\text{GNN}}(G;\theta_G)_i  
\end{equation}
between the input graph $G\in \mathcal{G}$ and the output labels $\hat{Y} \in \mathcal{Y}$, assigning to each node $i$ its predicted label $\hat{y}_i$. The training of GNN attempts to find parameters $\theta_G$ such that $\hat{y}_i \approx y_i$ on the training set. Note that standard GNNs are \textbf{\textit{graph-specific}} functions, i.e. $f_{\text{GNN}}(\cdot;\theta_G): G\mapsto \hat{Y}$, which do not generalize to other graphs, since other graphs $G' \in \mathcal{G}$ define distinct distributions of $Y'$, $\boldsymbol{A}'$,  and $\boldsymbol{X}'$, or even different types of features such as continuous, categorical, or text features. 

To solve the generalization problem mentioned above, this paper proposes to perform graph reasoning as a {\em text-to-text} problem~\citep{T5}, as shown in Figure~\ref{fig: GraphText} (b). Inspired by prompt tuning~\citep{GPT3,PromptSurvey}, we construct two graph-specific maps to form the input and output space of a text-to-text problem: a map $g: G \mapsto T_{in}$ that maps the graph input to text space, and a map $h: T_{out} \mapsto \tilde{Y}$ that maps the output of LLM to label predictions $\tilde{Y}$. In this way, we can use a generative large language model $f_{\text{LLM}}$ to perform graph reasoning as
\begin{equation}
\tilde{y}_i = h(f_{\text{LLM}}(g(G)_i;\phi))    
\end{equation}
where $g(G)_i=T_{in}[i]$ denotes the text sequence representing node $i$. Different from GNNs, $f_{\text{LLM}}(\cdot; \phi): \mathcal{T}\rightarrow\mathcal{T}$ is a \textbf{\textit{graph-shared}} 
function, where both input and output are in text space, i.e.\ $T_{in}, T_{out} \in \mathcal{T}$, which not only activates of parametric knowledge encoded in the model $f_{\text{LLM}}(\cdot; \phi)$, but also enables interactions between human and AI agents to facilitate graph reasoning.


Specifically, as node classification, link prediction and graph classification 
are essentially classification tasks, we can naturally formulate these graph reasoning tasks as multi-choice QA problem~\citep{MCP} and design $h$ as the map from predicted choice $T_{out}\in \mathcal{T}$ to the corresponding prediction $\tilde{Y}$. However, the design of $g$ that maps the structural graph information into the text space of natural language is still a non-trivial problem. 



The primary challenge in converting graph data to language lies in handling its relational structure, which fundamentally deviates from the one-dimensional sequential nature of text data. Inspired by linguistic syntax trees~\citep{MathematicalLogic07}, we introduce graph-syntax trees as a bridge between relational and sequential data. The traversal of such a tree produces a sentence in natural language, which is fed to LLM for graph reasoning. 
Specifically, as shown in Figure ~\ref{fig: GraphText} (c), we compose a graph-syntax tree consisting of node text attributes and inter-node relationships. Next, we describe how to compose the node text attributes and inter-node relationships in Section \ref{sec: graph info construction}, and how to build a graph-syntax tree in Section \ref{sec: graph info linearization}.

\subsection{Textual and Relational Information for Syntax Trees}
\label{sec: graph info construction}
A graph syntax tree is composed of both textual and relational information derived from the graph.
For textual information, \ours constructs a text attribute set $F \in \mathcal{T}$ for an arbitrary graph $G \in \mathcal{G}$ (with or without text-attributes) composed of multiple types of attributes for each node, e.g. feature and label, in natural language. Specifically, for each node $v_i$ and feature type $m$, we construct a text sequence $\mathrm{F}_m[i]$ in natural language:
\begin{equation}
    \mathrm{F}_m[i]=\{t_1, t_2, \cdots t_{l_m}\} ,\;\; \mathrm{F}_m[i]\in \mathcal{T},
\end{equation}
where the sequence is of length $l_m$.
Each text attribute $\mathrm{F}_m$ can be derived from either sequential text features or continuous features. For text features, they can be directly added to the text attributes $F$. For example, we can directly add the text sequences of ``title'' and ``abstract'' into $F$ for citation graphs. For continuous features, e.g.\ the raw feature $\boldsymbol{X}$ or other graph embeddings, we propose to use discretization methods, e.g.\ clustering, to transform the continuous feature into a discrete space and then derive sequential data from it. For simplicity, we use the cluster index of K-means to generate a sequence of length 1 for all continuous features as K-means is effective in our experiments.

For relational information, \ours derives a set of matrices $R$ where each $\boldsymbol{R}_n \in R$ is a $|V| \times |V|$ matrix, depicting one type of relationship between nodes. Choices of $\boldsymbol{R}_n$ may be the original graph~\citep{GCN}, high-order connectedness~\citep{JKNet}, page-rank matrices~\citep{PPNP}, or any matrices that encode node-pair information. These relationships play an important role in determining the nodes and structure of the graph-syntax tree, which further improves the graph text prompt.

\subsection{Graph-syntax Tree Composition}
\label{sec: graph info linearization}

We now describe how to build a graph prompt using a graph-syntax tree of the graph text attributes and relationships $F$ and $R$. 
%
By analogy to the syntax tree in linguistics, we define a graph-syntax tree as an ordered tree: a directed acyclic graph (DAG) with nodes $\tilde{T}\in \mathcal{T}$ and edges $\tilde{E}$. In a graph-syntax tree, e.g. the one in Figure \ref{fig: GraphText} (c), each node stores a \textit{text sequence} in natural language, where the root node is an empty node; the leaf nodes $\tilde{T}_\text{L}$ are text sequences in the graph text attributes, i.e. $\forall \,\mathrm{T}_i \in \tilde{T}_\text{L}, \mathrm{T}_i \in F$; the internal nodes $\tilde{T}_\text{I}$ are text sequences in natural language, i.e. $\forall \,\mathrm{T}_i \in \tilde{T}_\text{I}, \mathrm{T}_i \in \mathcal{T}$. A graph syntax tree is constructed in three steps: (1) construct an ego-subgraph~\citep{GraphSAGE} $G_i$ for target node $v_i$ based on relationship $R$ (2) select leaf nodes $\tilde{T}_\text{L}$ based on the relationship $R$. (3) build up internal nodes $\tilde{T}_\text{I}$ and edges $\tilde{E}$ based on the leaf nodes' types and their relationship with the graph\footnote{The hierarchy of the tree can be defined flexibly, but we have empirically discovered that a simple configuration, with attribute type at the top hierarchy and relation type at the bottom hierarchy for internal nodes, as illustrated in Figure ~\ref{fig: GraphText} (c), yields strong performance. Further details are available in Section \ref{sec: exp ablation}.}. Notably, the leaf nodes are sorted according to their relationships with the center-node, preserving the relative relationship in a one-dimensional order.


We illustrate this with a node classification example shown in Figure \ref{fig: GraphText} (c). Before building the graph-syntax tree, \ours determines the text attributes composed of raw features and observed labels, i.e. $F=\{ \mathrm{F}_X[i], \mathrm{F}_Y[i] \mid \forall v_i \in V \}$, and a relationship set composed of determined by shortest path distance (SPD): center-node, 1st-hop, and 2nd-hop, i.e. $R=\{\boldsymbol{R}_\text{SPD=0}, \boldsymbol{R}_\text{SPD=1}, \boldsymbol{R}_\text{SPD=2}\}$. Then, for target node $v_i$ (0 in the example), an ego-subgraph~\citep{GraphSAGE} (with nodes [0,1,2,3,4]) is sampled based on the relative relationship between $v_i$ and other nodes. Finally, a graph-syntax tree is constructed with leaf nodes $\tilde{T}_\text{L}=\{\mathrm{F}_\text{X}[0],\mathrm{F}_\text{X}[1],\mathrm{F}_\text{X}[2]\mathrm{F}_\text{X}[3],\mathrm{F}_\text{X}[4],\mathrm{F}_\text{Y}[1],\mathrm{F}_\text{Y}[3]\}$, the internal nodes $\tilde{T}_\text{I}=$\{``center-node'', ``1st-hop'', ``2nd-hop'', ``label'', ``feature''\}, and the corresponding edges. 
The traversal of the resulting graph-syntax tree leads to a text sequence in natural language.

Compared with the direct flattening of (sub)graphs~\citep{NLGraph, ExploreGraphLLM}, using a graph-syntax tree-based prompt has many advantages: Above all, unlike a graph, which has no topology order, a syntax tree is a DAG that can be topologically sorted, which gracefully converts a relational structure to a sequence of nodes. Moreover, \ours easily incorporates the inductive biases of GNNs through the construction of node text attributes $F$ and relationships $R$. For example, we can easily encode the feature-propagation mechanism of GNNs by including a text attribute derived from the propagated feature $\boldsymbol{A}^{k}\boldsymbol{X}$~\citep{GAMLP}, into the node attributes $F$. We can also incorporate the feature similarity-based aggregation~\citep{GAT} by adding $\boldsymbol{X}\boldsymbol{X}^{\top}$ to $R$. These graph-based inductive biases can significantly boost LLMs' graph reasoning performance (further discussed in Section~\ref{sec: exp training-free graph reasoning}). Last but not least, a tree naturally defines a hierarchical structure, which LLMs are proficient in reasoning on~\citep{CodeAsPolicies}, by training on code data~\citep{Codex} and web page data~\citep{LLaMA2}.

\section{Related Work}
\label{sec: related_work}
\paragraph{Unlock Graph Space for Language Models.} 
Large Language Models (LLMs)~\citep{GPT3,ChatGPT,PaLM2,GPT4} possess impressive reasoning capabilities~\citep{CoT,ToT,ComplexCoT}. At the heart of LLMs' reasoning prowess is their ability to process and generate natural language inputs and outputs, enabling flexible interactions~\citep{LMCascade} with both humans and AI agents. This unique capability empowers them with remarkable abilities such as complex reasoning~\citep{ComplexCoT} and decision-making~\citep{ReAct, CodeAsPolicies}. Despite their success, applying LLMs to relational graph data remains challenging, primarily due to the absence of a natural language representation for graphs. \ours bridges this gap by providing a novel framework that enables LLMs to seamlessly integrate and reason over relational graph data using the same natural language capabilities, thereby unlocking their potential for a wide range of graph-based applications.
\paragraph{Training-free Graph Reasoning} Graph neural networks (GNNs)~\citep{GCN,GIN} excel in handling relational graph data, thanks to the message-passing mechanism for aggregation and transformation of neighborhood representations. Their standout performance can be attributed to their intrinsic capability to assimilate graph inductive biases. This incorporation of biases is achieved by designing representations with the graph structure in perspective, such as position embeddings~\citep{LSPE,Graphormer,SAN} and propagated features~\citep{SGC,GAMLP}. Furthermore, they can introduce diverse aggregation methods, like feature similarity-based message passing~\citep{GAT,GAuG} or high-order aggregation~\citep{PPNP,PPRGo,GPRGNN}. However, as highlighted in Section \ref{sec: model framework}, due to the variance in both structure and feature, the majority of GNNs are \textit{graph-specific}. They are tailored for a particular graph type with consistent features and structures, thus posing challenges for generalization to different graphs.

In a parallel vein, \ours also taps into the potent ability to infuse graph inductive biases for graph reasoning, achieved through designing both the textual and relational aspects of the graph-syntax tree. Setting itself apart from GNNs, \ours approaches graph reasoning in a \textit{graph-shared} domain, facilitating the broader applicability of a single LLM to diverse graphs and offering training-free and interactive graph reasoning.

\paragraph{Connecting Both Worlds.} Recent endeavors~\citep{GIANT,GLEM,TAPE} have aimed to merge the language and graph domains. Most methods involve transitioning the problem into a graph-specific realm, utilizing a combination of a text-encoder (either pre-trained~\citep{GIANT} or learned~\citep{AdsGNN}) and a GNN predictor. This methodology still falls into a \textit{graph-specific} paradigm. Very recently, there are concurrent works~\citep{GPT4Graph,InstructGLM,NLGraph,ExploreGraphLLM} exploring to leverage LLMs for graph-related tasks. These methods either directly flatten the nodes and edges~\citep{GPT4Graph,NLGraph} or employ rule-based prompts on text-attributed graphs~\citep{ExploreGraphLLM,InstructGLM}.

Nevertheless, \ours is fundamentally different from these works. Foremost, \ours proposes a language defined by a graph-syntax tree, offering a flexible and structured approach for seamlessly integrating graph inductive biases.  
Moreover, it also serves as a general framework for graph reasoning, which can be applied to scenarios encompassing in-context learning and instruction tuning. It accommodates various types of graphs, including general graphs and text-attributed graphs, and is adaptable to both closed-source Large Language Models (LLMs) ~\citep{ChatGPT,GPT4} and open-source LLMs~\citep{LLaMA2}.
\begin{table}[!ht]
\caption{Node classification results (accuracy \%).
}
\label{tab: exp-icl}
  \centering
  \resizebox{\linewidth}{!}{
  \begin{tabular}{cccccccc}
  \toprule
  Model & Graph Text Attributes & Relations & \multicolumn{1}{c}{Cora} & \multicolumn{1}{c}{Citeseer} & \multicolumn{1}{c}{Texas} & \multicolumn{1}{c}{Wisconsin} & \multicolumn{1}{c}{Cornell} \\ \midrule
  GCN & NA & original & \underline{81.4} & \underline{69.8} & 59.5 & 49.0 & 37.8\\
  GAT & NA & original & 80.8 & 69.4 & 54.1 & 49.0 & 45.9 \\
  GCNII & NA & original & 81.2 & 69.8 & 56.8 & 51.0 & 40.5 \\
  GATv2 & NA & original & \textbf{82.3} & \textbf{69.9} & 62.2 & 52.9 & 43.2 \\
  \midrule
  \multirow{6}{*}{\textbf{\ours-ICL}}
  & \multirow{2}{*}{label} & original & 26.3  & 13.7 & 5.4 & 9.8 & 21.6 \\
  & \textbf{} &  ori.+synth. & 53.0  & 37.2 & 70.3 & \textbf{64.7} & \textbf{51.4} \\
  & \multirow{2}{*}{label+feat} & original  & 33.4 & 36.9 & 5.4 & 29.4 & 24.3 \\
  & \textbf{} & ori.+synth.  & 52.1 & 50.4 & \underline{73.0} & \underline{60.8} & \underline{46.0} \\
  & \multirow{2}{*}{label+feat+synth.} & original & 64.5  & 51.0 & \underline{73.0} & 35.3 & 48.7 \\
  & \textbf{} & ori.+synth. & 68.3  & 58.6 & \textbf{75.7} & 54.9 & \textbf{51.4} \\ \bottomrule
  \end{tabular}
}
\end{table}

\begin{figure*}[t]
    \centering
    \includegraphics[width=1\linewidth]{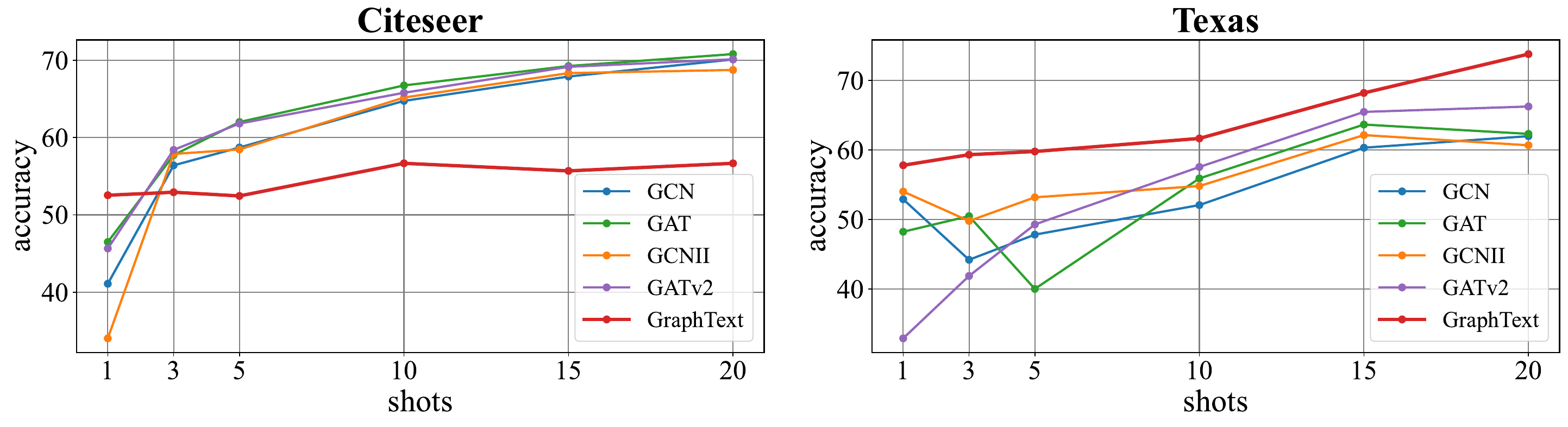}
    \caption{Few-shot in-context learning node classification accuracy. We perform 1, 3, 5, 10, 15, and 20-shot node classification on Citeseer and Texas datasets.}
    \label{fig: n_shots}
\end{figure*}
\section{Experiments}
We conduct extensive experiments to demonstrate the effectiveness of \ours. Firstly, in Section~\ref{sec: exp training-free graph reasoning}, we delve into the remarkable capacity of \ours for training-free graph reasoning. Subsequently, Section~\ref{sec: exp interactive} highlights the interactive graph reasoning capabilities of \ours. We further analyze various ablations of graph-syntax trees in Section~\ref{sec: exp ablation}. Concluding our exploration, Section~\ref{sec: exp tag-instruction-tuning} illustrates how \ours can seamlessly function as a versatile framework, catering to both in-context learning and instruction tuning across on both general graph and text-attributed graphs.

\subsection{Training-free Graph Reasoning}
\label{sec: exp training-free graph reasoning}

One unique ability of \ours is the training-free graph reasoning by in-context learning (ICL). In this section, we demonstrate this capability on the node classification tasks. Specifically, we use two citation datasets ({\em Cora} \citep{mccallum2000automating} and {\em Citeseer} \citep{giles1998citeseer}), and three webpage datasets ({\em Texas}, {\em Wisconsin}, and {\em Cornell}~\citep{GeomGCN}). The detailed discussion of experimental settings and the dataset statistics can be found in Appendices \ref{apx: exp-settings details} and \ref{apx: exp-settings data}.

We selected standard GNNs, including {\em GCN}~\citep{GCN} and {\em GAT}~\citep{GAT}, along with their more recent variants {\em  GCNII}~\citep{GCNII} and {\em GATv2}~\citep{GATv2}, as our baselines. These GNN baselines are supervised and specific to individual graphs, trained solely for inference on one dataset. In contrast, \ours utilizes a single pre-trained LLM (ChatGPT) for all datasets without any graph-specific training.

Given an input graph, \ours constructs a graph-syntax tree by incorporating two types of information: {\em text attributes} and {\em relations}. We utilize three types of text attributes for the graph: observed labels (referred to as {\em `label'}), features generated by K-means clustering of $\boldsymbol{X}$ (referred to as {\em `feat'}), and synthetic text (referred to as {\em `synth.'}) derived from feature- and label- propagation. Additionally, we employ two types of relations: the original graph structure ($\boldsymbol{A}$ with self-loops) and synthetic relations based on feature similarity, shortest-path distance, and personalized pagerank~\citep{PPNP}. The hyperparameter details are provided in Appendix \ref{apx: hyperpara}.

Experimental results are depicted in Table~\ref{tab: exp-icl} and Figure~\ref{fig: n_shots}. We observe that directly flattening raw text (labels) using relations, as proposed in~\citep{ExploreGraphLLM}, results in poor performance, occasionally worse than random. Incorporating discretized features into text attributes improves the performance slightly.

Integrating the graph inductive bias into both text attributes and relations enhances performance. The addition of synthetic relationships significantly boosts performance across all datasets, indicating that the raw graph lacks sufficient information and requires augmentation. This observation aligns with findings in graph structure learning literature~\citep{LDS,GAuG}. Furthermore, the inclusion of synthetic text attributes is beneficial in most cases. Ultimately, the combination of synthetic text attributes and synthetic relations yields the highest accuracy for \ours in four out of five datasets.

Remarkably, even though it is not trained on graph data, \ours surpasses several GNN baselines, particularly when the label rate is low (see Figure \ref{fig: n_shots}) and in heterophilic datasets. This is due to the merits that, in contrast with standard GNNs, \ours decouples depth and scope in graph reasoning~\citep{ZengZXSMKPJC21}. The strong performance of \ours in training-free graph reasoning highlights the substantial potential of leveraging LLMs in graph machine learning.

\begin{table}
    \label{tab: interact}

    \centering
    \caption{Interactive graph reasoning results (accuracy \%) on Cora (node \# 2188). The table showcases the performance of GPT-4 and ChatGPT before and after human interactions with 15 times of evaluation. The reasoning metrics include PPR, Center-node, and instances where the model was Confused to respond or Refused (Conf./Ref.) to make their reasoning/prediction. See Figure~\ref{fig: InteractWithHumanExample} (c) for details.}

    \begin{tabular}{llccccc}
        \toprule
        \multirow{2}{*}{Model} & \multirow{2}{*}{Interaction} &  \multirow{2}{*}{Accuracy} & \multicolumn{3}{c}{Reasoning} \\
        \cmidrule(lr){4-6}
        & & & PPR & Center-node & Conf./Ref. \\
        \midrule
        \multirow{2}{*}{GPT-4} & Before & 73.3 & 73.3 & 26.7 & 0 \\
        & After & \textbf{100 (+26.7)} & 100 & 0 & 0 \\
        \midrule
        \multirow{2}{*}{ChatGPT} & Before & 26.7 & 26.7 & 53.3 & 20.0 \\
        & After & \textbf{63.6 (+36.9)} & 72.7 & 18.2 & 9.1 \\
        \bottomrule
    \end{tabular}
\end{table}

\subsection{Interpretable and Interactive Graph Reasoning}
\label{sec: exp interactive}
In this section, we illustrate that \ours facilitates effective \textbf{interactive graph reasoning}: through its ability to generate and \textbf{clarify} predictions in natural language, both humans and LLMs can directly interact with \ours. 

To illustrate this concept, we will use Cora node \#2188. 
Figure~\ref{fig: InteractWithHumanExample} (a) shows two types of text attributes we use: the center-node pseudo labels and the PPR (Personalized PageRank) pseudo label sequence, where the first PPR neighbor denotes the most important label prediction. Upon examining the demonstrations (marked in blue), it becomes apparent that the PPR pseudo-labels provide a more robust mechanism for paper topic prediction. Utilizing either a count of PPR labels followed by a majority vote, or merely referencing the foremost PPR label, consistently results in the correct categorization in the given examples. Hence, based on these graph inductive biases derived from samples, we can reasonably figure out correct topic of the target paper should be A, which not only is the first entry, but also the predominant label in the PPR pseudo-label sequence.

We leverage \ours with ChatGPT and GPT-4 to perform graph reasoning on the provided example. Their respective reasoning processes and outcomes are illustrated and summarized in Figure~\ref{fig: InteractWithHumanExample} and Table~\ref{tab: interact} respectively, from which we draw several key insights:

\textbf{1. LLMs inherently possess the knowledge and inductive bias toward graph reasoning.} Specifically, both ChatGPT and GPT-4 acknowledge the importance of center-nodes and sometimes make predictions based on center-node labels. ChatGPT exhibits reasoning with the center-node bias 53.3\% of the time, while GPT-4 does so at a rate of 26.7\%. 

\textbf{2. LLMs can adjust their prior inductive bias based on demonstrations.} Through in-context learning, \ours can recalibrate their bias and make more accurate predictions. Our observations indicate that GPT-4 significantly outperforms ChatGPT, achieving an accuracy of 73.3\%, markedly superior to ChatGPT's 26.7\%.

\textbf{3. LLMs can adapt their prior inductive bias based on human feedback.} Figure~\ref{fig: InteractWithHumanExample} (b) provides an illustrative example, with a detailed reasoning of LLM can be found in Appendix \ref{apx: examples InteractionWithHuman}. Specifically, after human interaction, GPT-4 shows remarkable adaptability, achieving an impeccable accuracy of 100\% and adhering to the PPR logic. Meanwhile, ChatGPT also enhances its performance notably (gaining 36.9\% in accuracy), but occasionally maintains its antecedent biases.

In summary, through graph reasoning in natural language, \ours can effectively leverage its pre-trained knowledge to engage in graph reasoning and, crucially, adapt its existing knowledge through demonstrations or external feedback.

\begin{figure*}[!t]
    \centering
    \includegraphics[width=0.99\linewidth]{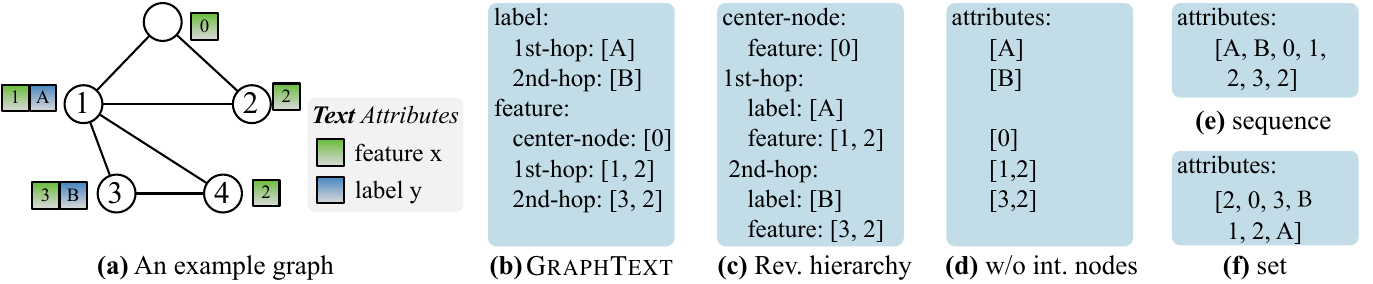}
    \caption{Ablations of graph-syntax trees. \textbf{(a)} An example graph. \textbf{(b)} \ours text prompt (The full example can be found in Figure ~\ref{fig: GraphText}). \textbf{(c-f)} Text prompts of different tree designs.}
\label{fig: TreeAblation}
\end{figure*}

\begin{table}[!t]
\centering
\caption{Ablations of \ours on Cora, Citeseer and Texas.}
\label{table: TreeAblation}
\begin{tabular}{lcccccc}
\toprule
\multirow{2}{*}{Model} & \multicolumn{2}{c}{Cora} & \multicolumn{2}{c}{Citeseer} & \multicolumn{2}{c}{Texas} \\
\cmidrule(lr){2-3} \cmidrule(lr){4-5} \cmidrule(lr){6-7}
& Acc. \% & $\Delta$ & Acc. \% & $\Delta$  & Acc. \% & $\Delta$ \\
\midrule
\textbf{\ours} & \textbf{76.5} & - & \textbf{58.6} & - & \textbf{75.7} & - \\
\quad rev. hierarchy & 68.3 & -10.7 \% & 57.6 & -1.7 \% & 73.0 & -3.6  \%\\
\quad w/o int. nodes & 67.8 & -11.4 \% & 56.3 & -3.9 \% & 75.7 & -0  \%\\
\quad sequence & 67.0 & -12.4  \% & 53.0 & -9.6 \% & 70.3 & -7.1  \%\\
\quad set & 65.9 & -13.9 \% & 56.4 & -3.8 \% & 67.6 & -10.6 \% \\
\bottomrule
\end{tabular}
\end{table}


\subsection{Ablation Studies on Graph-syntax Trees}
\label{sec: exp ablation}
The graph-syntax tree serves as the core design of \ours, transforming a graph into a 1-dimensional natural language sequence. Within \ours, the text and relational data of a graph are initially formulated, followed by the construction of a graph-syntax tree. This section delves into the ablations of various methods for building graph-syntax trees.

As shown in Figure~\ref{fig: TreeAblation}, besides the proposed \textit{\ours} method for constructing a graph-syntax tree, we present four ablation types: (1) Reverse hierarchy (denoted as \textbf{rev. hierarchy} in Figure~\ref{fig: TreeAblation} (c): The tree hierarchy is inverted, positioning the relationship type at the top and the text attribute type at the bottom. (2) Without internal nodes (denoted as \textbf{w/o int. nodes} in Figure~\ref{fig: TreeAblation} (d)): The internal nodes of the graph-syntax tree are eliminated, but the \textit{\ours} hierarchy remains intact (note the indents are kept, maintaining the hierarchical structure of the tree). (3) Sequential prompt (denoted as \textbf{sequence} in Figure~\ref{fig: TreeAblation} (e)): The tree hierarchy is removed, yielding a sequence of text attributes. (4) Set prompt (denoted as \textbf{set} in Figure~\ref{fig: TreeAblation} (f)): Sequence order is removed, yielding a set.

From Table~\ref{table: TreeAblation}, several observations can be made:
(1) The graph-syntax tree of \ours consistently outperforms the others, underscoring the efficacy of our approach.
(2) The hierarchical structure of the tree plays a crucial role in the design of the graph prompt. Specifically, we observe a sheer performance drop when using a \textit{sequence} or a \textit{set} to represent the graph information. Upon inspecting the LLM's reasoning, we found it treats graph learning purely as label counting without recognizing structure.
(3) Variations in the tree hierarchy design can impact performance; for instance, \textit{rev. hierarchy} underperforms compared to \ours.
(4) The comparison between \textit{w/o int. nodes} and \ours reveals the importance of making LLMs aware of text attribute types. The only exception is the Texas dataset since all types of attributes are almost identical in this dataset (detailedly discussed in Appendix~\ref{apx: examples texas}). This suggests that LLMs utilize text descriptions to distinguish and understand different attributes during graph reasoning.

\subsection{Experiments on Text-attributed Graph}
\label{sec: exp tag-instruction-tuning}
\begin{table}[h]
\centering
\caption{Node classification results (accuracy \%) on real-world text attributed graphs. Experiments are conducted using in-context learning with ChatGPT, as well as instruction tuning with Llama-2-7B. Note that ``text'' refers to raw text attributes, while ``feat'' represents the continuous features on the graph. The top results for each category are highlighted in bold.}
\label{tab: exp-sft}
\begin{tabular}{llcc}
\toprule
Framework & Model & Cora & Citeseer \\
\midrule
\multirow{2}{*}{GNNs} 
& GCN & 89.13 & 74.92 \\
& GAT & \textbf{89.68} & \textbf{75.39} \\
\midrule
\multirow{6}{*}{\ours} 
& ChatGPT-text & \textbf{67.77} & \textbf{68.98} \\
& ChatGPT-feat & 10.68 & 16.14 \\
& ChatGPT-feat+text & 65.19 & 66.46 \\ 
\cmidrule(lr){2-4}
& Llama-2-7B-text & 60.59 & 49.37 \\
& Llama-2-7B-feat & \textbf{87.11} & \textbf{74.77} \\
& Llama-2-7B-feat+text & 77.53 & 73.83 \\
\bottomrule
\end{tabular}
\end{table}





In this section, we demonstrate that \ours is also applicable to text-attributed graphs. As depicted in Table~\ref{tab: exp-sft}, we conducted training-free node classification on the Cora and Citeseer datasets with both raw text attributes~\citep{ExploreGraphLLM} and continuous features~\citep{Sentence-BERT}. We observed that using closed-source LLMs, such as ChatGPT, the performance lags behind the GNN baseline methods. Thus, we further explored the potential of instruction tuning on currently available open-source LLMs, such as Llama-2~\citep{LLaMA2}. For natural language prompts construction, we adopted an approach almost identical to the in-context learning setting. Furthermore, we expand the original vocabulary of Llama-2 by introducing selected options as new tokens and then fine-tune the large language model by the widely-used and efficient Low-Rank Adaptation (LoRA)~\citep{LoRA}.

From the results in Table 4, it is evident that even with a relatively smaller open-source model, Llama-2-7B, our best results from instruction tuning across various settings surpass those of ChatGPT and approach the GNN baselines. This validates that our method can be beneficial in an instruction-tuning scenario. It also implies that using \ours, we can feasibly fine-tune smaller open-source LLMs with reasonable computational costs, achieving performances that can rival or even surpass those of much larger closed-source models, such as ChatGPT or GPT-4.

Another intriguing observation is the notably poor performance of ChatGPT in settings incorporating continuous feature – nearing a random guess. This is attributable to the inherent limitation of these closed-source LLMs: they are designed to process raw discrete text inputs and fail to directly handle the continuous inputs. In contrast, open-source LLMs possess the ability to map these continuous embeddings into their embedding space, facilitating improved performance.

Upon contrasting these two groups of models, we noticed a decline in the performance of open-source models when processing raw text inputs. This decline can be ascribed to the constraints imposed by the size of the LLM parameters and the volume of pre-training corpora used. It suggests that harnessing larger-scale open-source models, such as Llama-2 variants including 13B, 30B, and 70B, would significantly bolster their modeling capacity for raw text. Concurrently, by leveraging the ability to process continuous embeddings, these models would inevitably exhibit enhanced graph reasoning capabilities, paving the way for more sophisticated graph-based applications.

\section{Conclusion}
In this paper, we propose \ours, a framework that enables graph reasoning in text space. It easily incorporates the inductive bias of GNNs by constructing a graph-syntax tree. The traversal of a graph-syntax tree leads to a graph prompt in natural language and is fed to LLM to perform graph reasoning as text generation. \ours enables training-free graph reasoning where a \ours-LLM can deliver performance on par with, or even surpassing, supervised graph neural networks through in-context learning. What's more, \ours fosters explainable and interactive graph reasoning: \ours performs graph reasoning in natural language which enables humans and LLMs to engage with graph learning using natural language. These abilities highlight the immense and largely untapped potential of LLMs in the realm of graph machine learning.

\section*{Ethics Statement}
Graphs are prevalent in the real world. On the bright side, \ours alleviates the computational load and carbon footprint associated with training numerous non-transferable, graph-specific models. However, while the training-free graph reasoning capability of \ours introduces minimal costs, there's potential for misuse in malicious recommendation systems and malware.

\bibliography{references.bib}
\bibliographystyle{iclr2024_conference}

\newpage

\appendix

\onecolumn

\section*{Reproducibility Statement}

The code to reproduce our results will be available soon. Our experimental settings and implementation details are stated in Section \ref{apx: exp-settings details}, and the important hyper-parameters are discussed in Appendix \ref{apx: hyperpara} section.

\section{Experimental Settings}

\subsection{Implementation Details}
\label{apx: exp-settings details}
In our experiments node classification is approached as a multi-choice QA challenge, and we employ the prompt detailed in Appendix ~\ref{apx: prompt}. The raw text attributes can be directly leveraged to form the textual information $F$. Additionally, we utilize the dataset metadata to extract raw text labels, creating a textual feature for each node. Nodes without data are assigned the value ``NA''. Consequently, every general graph can be perceived as a text-attributed graph with a minimum of one type of text attribute, namely the label. For continuous attributes, we use K-means to discretize continuous features (with the $K$ being the number of classes).

During the in-context-learning experiments, to prevent meaningless demonstrations lacking neighboring labels, we choose a sample with the highest degree from each label set. For the instruction tuning experiments using open-source LLMs, we can leverage continuous attributes in a more flexible way. Specifically, inspired by multi-modality LLMs~\cite{VisionLLM}, we use a Multi-Layer Perceptron (MLP) projector to map continuous features into the input text space, the token embedding space of LLaMA-2~\citep{LLaMA2}. We utilize AdamW~\citep{AdamW} in conjunction with DeepSpeed~\citep{DeepSpeed} to train the huggingface LLaMA2-7b model~\footnote{https://huggingface.co/meta-llama/Llama-2-7b-hf}, with FP16 activated.

\subsection{Datasets}
\label{apx: exp-settings data}

\begin{table*}[htbp] 
\caption{The statistics of the datasets.}
\label{tab: datasets}
\small
\begin{tabular}{c|ccccccc}
\toprule
Benchmarks  & \#Nodes & \#Edges & \#Classes & \#Features & \#Train      &\#Validation     &\#Test \\ \midrule
Cora        & 2708    & 5278    & 7         & 1433       & 140          & 500               &1000    \\ 
Cora-TAG    & 2708    & 5278    & 7         & 1433       & 1624          & 541               &543    \\
Citeseer    & 3327    & 4552    & 6         & 3703       & 120          & 500               &1000    \\
Citeseer-TAG    & 3327    & 4552    & 6         & 3703       & 1911          & 637                &638    \\
Cornell     & 183    & 298   & 5         & 1703        & 87          & 59               &37    \\
Texas     & 183    & 325   & 5         & 1703        & 87          & 59               &37    \\
Wisconsin & 251    & 515  & 5         & 1703       & 120          & 80              &51     \\ \bottomrule
\end{tabular}
\end{table*}

In this section, we provide more relevant details about the datasets we used in experiments. The dataset statistics are provided in Table \ref{tab: datasets}. The datasets can be categorized into citation network datasets (i.e. Cora, Citeseer, and ogbn-arxiv) and web-page networks (i.e. Cornell, Texas, and Wisconsin), additionally, we use two text-attributed-graph (TAG) version of Cora and Citeseer, denoted as Cora-TAG and Citeseer-TAG.

\textbf{Citation graphs}: Most GNN-related studies, as referenced in works like~\cite{GCN,GAT}, often employ citation networks as benchmarking tools. Within these networks, nodes represent papers from the computer science domain. The features of these nodes are derived from bag-of-word vectors of the respective paper titles. Edges depict the citation links between these papers, while labels indicate the paper's specific categories. The text attributes are the title and abstract of the paper.

\textbf{WebKB graphs}~\cite{GeomGCN}: Sourced by Carnegie Mellon University, aggregates web pages from computer science departments across several universities. We employ three specific subsets from this collection: Cornell, Texas, and Wisconsin. In these subsets, each node symbolizes a web page, while edges denote hyperlinks connecting them. Nodes are characterized by a bag-of-words representation derived from their respective web pages. These pages have been meticulously categorized into five distinct classes: student, project, course, staff, and faculty.

The datasets mentioned above can be found in the following URLs:
Cora~\footnote{https://relational.fit.cvut.cz/dataset/CORA}, Citeseer~\footnote{https://linqs.soe.ucsc.edu/data},
Cora-TAG~\footnote{https://github.com/CurryTang/Graph-LLM},
Citeseer-TAG~\footnote{https://github.com/CurryTang/Graph-LLM},
Texas~\footnote{https://docs.dgl.ai/en/0.8.x/api/python/dgl.data.html\#node-prediction-datasets},
Cornell~\footnote{https://docs.dgl.ai/en/0.8.x/api/python/dgl.data.html\#node-prediction-datasets},
Wisconsin~\footnote{https://docs.dgl.ai/en/0.8.x/api/python/dgl.data.html\#node-prediction-datasets}.

\subsection{Hyperparameters}
\label{apx: hyperpara}

In \ours, the selection of text attributes $F$ and relations $R$ are the most important parameters. Here we discuss their choices and report the selected parameters in Table \ref{tab: hyperparameters}.
For text attributes $F$, there are several choices: propagated features $\boldsymbol{A}^{k}\boldsymbol{X}$, propagated labels $\boldsymbol{A}^{k}\boldsymbol{Y}_L$, raw features $\boldsymbol{X}$ and labels $\boldsymbol{Y}_L$; For relationship $R$, there are several choices: k-hop shortest-path distance, denoted as $\boldsymbol{S}_{k}$, propagated feature similarity, denoted as $\text{sim}(\boldsymbol{A}^k\boldsymbol{X})$, and pagerank matrix~\cite{PPNP}, with restart probability $\alpha=0.25$, denoted as $\boldsymbol{\Pi}$.
\begin{table}[h]
\centering
\caption{\ours in-context learning hyperparameters.}
\label{tab: hyperparameters}
\begin{tabular}{l|cc}
\toprule
& \textbf{Text Attributes} & \textbf{Relations} \\
\midrule
\textbf{Cora} & $\boldsymbol{A}^2\boldsymbol{Y}_L$,$\boldsymbol{A}^3\boldsymbol{Y}_L$ & $\boldsymbol{S}_{0}$,$\boldsymbol{\Pi}$, $\text{sim}(\boldsymbol{A}^2\boldsymbol{X})$,$\text{sim}(\boldsymbol{A}^3\boldsymbol{X})$  \\
\textbf{Citeseer} & $\boldsymbol{X}$, $\boldsymbol{A}^3\boldsymbol{Y}_L$ & $\boldsymbol{S}_{0}$,$\boldsymbol{S}_{2}$,$\boldsymbol{\Pi}$, $\text{sim}(\boldsymbol{A}^2\boldsymbol{X})$ \\
\textbf{Taxes} & $\boldsymbol{A}^2\boldsymbol{Y}_L$ & $\boldsymbol{S}_{0}$,$\boldsymbol{S}_{2}$ \\
\textbf{Wisconsin} & $\boldsymbol{Y}_L$ & $\boldsymbol{S}_{0}$,$\text{sim}(\boldsymbol{X})$, $\text{sim}(\boldsymbol{A}\boldsymbol{X})$\\
\textbf{Cornell} & $\boldsymbol{A}^2\boldsymbol{Y}_L$ & $\boldsymbol{S}_{0}$, $\boldsymbol{S}_{0}$,$\text{sim}(\boldsymbol{X})$, $\text{sim}(\boldsymbol{A}\boldsymbol{X})$ \\
\bottomrule
\end{tabular}
\end{table}

\section{Prompt Examples}
\label{apx: prompt}
\subsection{Few-shot In-context Learning}
\begin{tcolorbox}[breakable, width=1\textwidth]
\textbf{Example of Citeseer:}\\
\textbf{[Human]}: You are a helpful assistant that classifies the topic of an academic paper based on the labels of the cited papers. You are going to choose the correct answer from several choices of paper categories: [A: Agents, B: Artificial Intelligence, C: Database, D: Information Retrieval, E: Machine Learning, F: Human Computer Interaction]\\

Here are a few examples:\\
\textless information\textgreater \\
\phantom{MM} \textless third-order\_pseudo\_labels\textgreater \\
\phantom{MM} \phantom{MM} \textless center\_node\textgreater ['A']\textless /center\_node\textgreater \\
\phantom{MM} \phantom{MM} \textless 1st\_feature\_similarity\_graph\textgreater ['A', 'A', 'A']\textless /1st\_feature\_similarity\_graph\textgreater \\
\phantom{MM} \phantom{MM} \textless ppr\textgreater ['A', 'B', 'A']\textless /ppr\textgreater \\
\phantom{MM} \textless /third-order\_pseudo\_labels\textgreater \\
\textless /information\textgreater \\
\textless question\textgreater What's the topic of academic paper given the information above?\textless /question\textgreater \\
\textless answer\textgreater A\textless /answer\textgreater \\
\\
Remaining examples \ldots \\
\\
Now let's answer the question below:\\
\textless information\textgreater \\
\phantom{MM} \textless third-order\_pseudo\_labels\textgreater \\
\phantom{MM} \phantom{MM} \textless 1st\_feature\_similarity\_graph\textgreater ['C', 'B', 'B']\textless /1st\_feature\_similarity\_graph\textgreater \\
\phantom{MM} \phantom{MM} \textless ppr\textgreater ['C']\textless /ppr\textgreater \\
\phantom{MM} \textless /third-order\_pseudo\_labels\textgreater \\
\textless /information\textgreater \\

What's the topic of the paper given the information above? Valid choices are [A: Agents, B: Artificial Intelligence, C: Database, D: Information Retrieval, E: Machine Learning, F: Human computer interaction].
Remember, your answer should be in the form of the class choice wrapped by \textless answer\textgreater  \textless /answer\textgreater .\\ \\
\textbf{[Assistant]}: \textless answer\textgreater C\textless /answer\textgreater

\label{icl prompt example}
\end{tcolorbox}

\subsection{Instruction Tuning}
\begin{tcolorbox}[breakable, width=1\textwidth]
\textbf{Example of Cora:}\\
\textbf{[Human]}: Your goal is to perform node classification. You are given the information of each node in a xml format. Using the given information of a node, you need to classify the node to several choices: [\textless c0\textgreater : Rule\_Learning, \textless c1\textgreater : Neural\_Networks, \textless c2\textgreater : Case\_Based, \textless c3\textgreater : Genetic\_Algorithms, \textless c4\textgreater : Theory, \textless c5\textgreater : Reinforcement\_Learning, \textless c6\textgreater : Probabilistic\_Methods]. Remember, your answer should be in the form of the class label. \\
\textless information\textgreater \\ 
\phantom{MM} \textless feature\textgreater \\
\phantom{MM}\phantom{MM} \textless center\_node\textgreater \textless x\textgreater \textless x emb\textgreater \textless /x\textgreater \textless /center\_node\textgreater \\
\phantom{MM}\phantom{MM} \textless 1st\_feature\_similarity\_graph\textgreater \textless x\textgreater \textless x emb\textgreater \textless /x\textgreater \textless /1st\_feature\_similarity\_graph\textgreater \\ 
\phantom{MM} \textless /feature\textgreater \\ 
\textless /information\textgreater \\ \\
\textbf{[Assistant]}: The answer is: \textless c6\textgreater 

\label{sft prompt exaple}
\end{tcolorbox}
Note that the ``$<$x emb$>$'' is the text token embedding for feature feature ``x'' generated by the MLP projector discussed in Appendix~\ref{apx: exp-settings details}.
\subsection{Examples of Texas}

\label{apx: examples texas}
\begin{tcolorbox}[breakable, width=1\textwidth]
\textbf{Node \# 132, 136, 143, \ldots} \\
Graph information: \\
pseudo labels: \\
center-node:['D'] \\
second-hop neighbor:['D', 'D', 'D', 'D', 'D'] \\
Target class: D \\

\textbf{Node \# 30} \\
Graph information: \\
pseudo labels: \\
center-node:['D'] \\
second-hop neighbor:['D', 'E', 'D', 'D', 'D'] \\
Target class: D \\

\textbf{Node \# 158} \\
Graph information: \\
pseudo labels: \\
center-node:['A'] \\
second-hop neighbor:['A'] \\
Target class: A \\

\end{tcolorbox}

We can observe that for the best setting in the Texas datasets, with hyperparameters discussed in Table~\ref{tab: hyperparameters}, the center-node pseudo labels mostly assemble the second-hop neighbors. Consequently, removing the text information, i.e. removing the internal nodes in the graph-syntax tree in Section \ref{sec: exp ablation} does not hurt the performance.

This also shows the advantage of decoupling depth and scope~\cite{ZengZXSMKPJC21} in the graph-syntax tree of \ours, which explains the performance gain of \ours over standard GNNs, e.g. GCN and GAT. A similar observation is also drawn in Figure 8 (i) of ~\citep{GPRGNN}, where $\boldsymbol{A}^2$ serves as the most important high-order aggregation scheme for Texas dataset.

\section{Interactive Graph Reasoning}
\label{apx: examples InteractionWithHuman}
Since \ours facilitates graph learning within a textual domain, it allows for direct interaction between both humans and AI agents. In this section, we spotlight the interactive graph reasoning capabilities of \ours using a practical example. First, we demonstrate how \ours can engage in self-interaction via zero-shot chain of thought reasoning. Following that, we illustrate how human interactions can guide \ours to refine its graph reasoning approach.

\subsection{Zero-shot Chain of Thought Reasoning}
\label{apx: interactive reasoning - zero shot}
Below is the example of graph reasoning on Cora node \#2188 in the setting of standard zero-shot chain of thought reasoning~\cite{CoT}\footnote{ChatGPT and GPT-4 perform reasoning with explanation by default. Therefore, there is no necessity to add the additional ``Let's think step-by-step'' prompt for the text input.} The input prompt for ``Cora node \#2188'' is as below:

\begin{tcolorbox}[breakable, width=1\textwidth]

\textbf{Input Prompt Cora node \#2188:}\\
\textbf{[Human]}: You are a helpful assistant that generates a classifies the topic of an academic paper based on the labels of the cited papers. You are going to choose the correct answer from several choices of paper categories:[A: Theory, B: Reinforcement Learning, C: Genetic Algorithm, D: Neural Network, E: Probabilistic Method, F: Case Based, G: Rule Learning] \\
Here are a few examples: \\
Graph information: \\
pseudo labels: \\
center-node:['F'] \\
ppr:['A', 'A', 'A', 'A'] \\
Topic of paper: A \\ \\
Graph information: \\
pseudo labels: \\
center-node:['B'] \\
ppr:['B', 'B', 'B', 'B'] \\
Topic of paper: B \\ \\
Graph information: \\
pseudo labels: \\
center-node:['C'] \\
ppr:['C', 'C', 'C', 'C'] \\
Topic of paper: C \\ \\
Graph information: \\
pseudo labels: \\
center-node:['D'] \\
ppr:['D', 'D', 'E', 'D'] \\
Topic of paper: D \\ \\
Graph information: \\
pseudo labels: \\
center-node:['A'] \\
ppr:['E', 'E', 'A', 'A'] \\
Topic of paper: E \\ \\
Graph information: \\
pseudo labels: \\
center-node:['F'] \\
ppr:['F', 'F', 'F', 'F'] \\
Topic of paper: F \\ \\
Graph information: \\
pseudo labels: \\
center-node:['G'] \\
ppr:['G', 'G', 'G', 'G'] \\
Topic of paper: G \\ \\
Now let's answer the question below: \\
Graph information: \\
pseudo labels: \\
center-node:['G'] \\
ppr:['A', 'G', 'A', 'E'] \\ 
What's the topic of the paper given the information above? Valid choices are [A: Theory, B: Reinforcement Learning, C: Genetic Algorithm, D: Neural Network, E: Probabilistic Method, F: Case Based, G: Rule Learning]
\end{tcolorbox}
As discussed in Section \ref{sec: exp interactive}, according to the demonstrations, the PPR pseudo-labels provide a more robust mechanism for paper topic prediction. Utilizing either a count of PPR labels followed by a majority vote, or merely referencing the foremost PPR label, consistently results in the correct categorization in the given examples. Hence, based on these graph inductive biases derived from samples, we can reasonably figure out correct topic of the target paper should be A, which not only is the first entry, but also the predominant label in the PPR pseudo-label sequence. Now, let's use ChatGPT and GPT-4 to answer the problem. We performed 15 times of evaluation and here we show one sample of correct and incorrect responses below:
\begin{tcolorbox}[breakable, width=1\textwidth]
\textbf{Correct Response without Human Interaction}\\
\textbf{[ChatGPT]}: Among the PPR labels, 'A' appears twice, 'G' appears once, and 'E' appears once. 'A' is \textcolor{mygreen}{the most frequent} label, so we classify the topic of the paper as 'A.'

Therefore, the topic of the paper is A: Theory. \\

\textbf{[GPT-4]}: Given the examples provided:

1. Whenever the center-node label is consistent with the majority of the PPR pseudo labels, the topic of the paper matches the center-node label.\\
2. However, when there's a discrepancy between the center-node label and the majority of the PPR labels, the topic of the paper is typically determined by \textcolor{myblue}{ the first (most crucial) label in the PPR list}.
\\
Given the graph information provided:
\\
- The center node is labeled 'G' which corresponds to 'Rule Learning'.\\
- The 'ppr' nodes are labeled 'A', 'G', 'A', and 'E', which corresponds to 'Theory', 'Rule Learning', 'Theory', and 'Probabilistic Method' respectively.
\\
It seems that while the center node is 'G', the cited papers are mostly of category 'A' and 'E'.
\\
From the pattern observed in the examples, the category that has the majority or at least equal emphasis with other categories in the 'ppr' nodes tends to be the category of the paper.

Therefore, in this case, 'A' or 'Theory' has the highest count in the 'ppr' nodes.

Thus, the topic of the paper given the information above is: A: Theory.
\end{tcolorbox}

\begin{tcolorbox}[breakable, width=1\textwidth]
\textbf{Wrong Response without Human Interaction}
\\
\textbf{[ChatGPT]}: Based on the labels in the "ppr" list, we can see that 'A' (Theory), 'G' (Rule Learning), and 'E' (Probabilistic Method) are mentioned. However, \textcolor{myred}{the 'center-node' label} is 'G' (Rule Learning), which indicates a stronger association with "Rule Learning." So, the topic of the paper is most likely "Rule Learning (G)." \\

\textbf{[GPT-4]}: Given the graph information: pseudo labels: center-node:['G'] ppr:['A', 'G', 'A', 'E'] \\
The center node label is 'G', which corresponds to "Rule Learning". The papers referenced (ppr) are mixed between "Theory", "Rule Learning", and "Probabilistic Method". However, since \textcolor{myred}{the center node holds the primary topic label}, the academic paper's topic is: G: Rule Learning. 
\end{tcolorbox}
From the analysis above, it's clear that:

By recasting the graph learning challenge into problems that both humans and LLMs can interpret and reason with, \ours facilitates LLMs in graph reasoning. For instance, ChatGPT can deduce the right answer through label counting, as highlighted in \textcolor{mygreen}{green}. Impressively, GPT-4 discerns the correct solution by employing the Personalized PageRank (PPR) prior, as illustrated in \textcolor{myblue}{blue}, and identifies the first PPR neighbor as the most important factor for prediction.

However, the lucid explanations rendered in natural language allowed us to pinpoint certain inadequacies in their graph reasoning approach:

\textbf{Failure to Retrieve Prior Knowledge}: Table~\ref{tab: interact} shows that LLMs, particularly ChatGPT, don't always effectively recall their previous understanding of graphs; only 26.7\% of ChatGPT's reasoning relies on PPR.

\textbf{Persistence in Prior Belief}: Despite the inherent reasoning prowess of LLMs, they can sometimes be hindered by preconceived notions about graphs. A detailed examination of the results in Table~\ref{tab: interact}, accompanied by the explanations of reasoning above, indicates that a deep-seated bias - where the central node is viewed as the most vital (denoted in \textcolor{myred}{red}) - is a primary cause of mistakes.

In the following section, we engage with \textit{\ours} to enhance its graph reasoning capabilities.

\subsection{Graph Reasoning with Human Interaction}

As highlighted in the preceding section, LLMs occasionally struggle with recalling prior knowledge and remain anchored to their pre-existing beliefs. To address this, we engage with \ours by: (1) providing a clear definition of PPR, and (2) explicitly instructing them to assess the reliability of both center-node pseudo and PPR pseudo labels. This leads to the following message:

\begin{tcolorbox}[breakable, width=1\textwidth]
\textbf{Human Interaction}
Note that PPR pseudo labels represent a sequence of labels ranked by importance, with the first one being the most crucial. Let's evaluate the reliability of center node pseudo labels against PPR pseudo labels and then make another prediction.
\end{tcolorbox}

With the interaction above, in most cases LLMs refined their answer, below are some examples:

\begin{tcolorbox}[breakable, width=1\textwidth]
\textbf{Correct Reasoning after Human Interaction}
\\
\textbf{[ChatGPT]}: In this PPR list, the labels 'A' and 'G' are present, with 'A' appearing before 'G,' which means 'A' is considered more crucial in the context of this paper. Now, let's make a prediction based on this evaluation: Since 'A' is the most crucial label in the PPR list, we will prioritize it for classifying the topic of the paper. Therefore, the paper's primary topic is "A: Theory" (Category A). The presence of 'G' in the PPR list indicates some relevance to "Rule Learning" (Category G), but because it comes after 'A,' it is considered less important in this context. So, the revised prediction is that the primary topic of the paper is "A: Theory." \\

\textbf{[GPT-4]}: Given the new information that PPR pseudo labels represent a sequence of labels ranked by importance, let's re-evaluate the provided data: \\
The center node suggests the paper is about 'G' (Rule Learning). However, considering the prominence of 'A' (Theory) in the ranked sequence, and since it's the most crucial label, the paper might be primarily focused on theoretical aspects related to Rule Learning. Considering both the center node and the importance sequence, the topic of the paper should be: A: Theory \\
\textbf{\textcolor{myblue}{Refusion} after Human Interaction}
\\
\textbf{[ChatGPT]}: Certainly, let's evaluate the reliability of center node pseudo labels against PPR pseudo labels.

In the provided graph information: The center-node label is 'G,' which suggests that the most important label for this paper is 'G.' However, when we consider the PPR labels, we see that 'A' appears as the first and third label, which might indicate some importance as well.

Let's consider both the center-node and the PPR labels for our prediction. Given that 'G' is the center-node label and 'A' appears in the PPR labels, we can prioritize the center-node label but also consider the presence of 'A' in the PPR labels.

So, the prediction for the topic of the paper can be: 
Topic of paper: G (with some influence from A).\\
\textbf{\textcolor{myred}{Confusion} after Human Interaction}
\\
\textbf{[ChatGPT]}: Based on this information, we can classify the topic of the paper as a combination of the most frequent labels among the neighboring papers, which are 'A' and 'G'. Therefore, the topic of the paper is likely a combination of topics 'A' and 'G'. \\
So, the answer is a combination of two topics, and the closest match among the given options is: \\
Option: F: Case Based \textit{\# Note that F does not present in the pseudo labels of the question}
\end{tcolorbox}

The provided examples, along with the consolidated findings in Table~\ref{tab: interact}, compellingly show that both ChatGPT and GPT-4 can adjust their pre-existing biases about graphs when given human feedback. Notably, after this interaction, GPT-4 delivers a flawless accuracy rate of 100\%, consistently following the PPR logic. Meanwhile, ChatGPT also sees a significant performance boost, with an accuracy improvement of 36.9\%. However, as evidenced in the examples, ChatGPT occasionally \textcolor{myblue}{refuses} updating its predictions or becomes \textcolor{myred}{confused}.

\section{Limitations and Future Work}

While the \ours framework offers notable advantages, there is ample room for enhancement and exploration of new applications.

One primary concern not fully addressed in this paper is the question of \textit{how to discretize continuous features}. As evident from Table~\ref{tab: hyperparameters}, most optimal settings are label-based. This makes \ours resemble a label propagation model, with Citeseer being the only exception. We posit that the observed trend might be attributed to two main factors: 1) The ineffectiveness of the discretization process, and 2) The discord between feature and label spaces, making reasoning challenging for LLMs.

Additionally, the \textit{design space of the graph-proxy tree is extensive} and often requires either expert knowledge or hyperparameter optimization. Although \ours boasts flexibility and generality, crafting the text-attribute set $F$ and relation set $R$, and determining their combinations result in a vast search space. However, given that \ours operates on a training-free paradigm, hyperparameter optimization can be swift.

Notwithstanding its constraints, \ours introduces avenues for novel research. Chiefly, it sets the stage for graph reasoning in natural language. Emerging advancements in the LLM realm can potentially be integrated into the graph ML domain. This includes areas such as multi-step reasoning~\citep{ToT}, decision-making~\citep{ReAct, CodeAsPolicies}, tool utilization~\citep{Toolformer}, and multi-agent collaboration~\citep{GenerativeAgents, MetaGPT}. Furthermore, the prospect of training-free graph learning streamlines the validation process for graph model designs. If one assumes the optimal relation set $R$ and feature set $H$ to be transferable, significant training time can be saved. Researchers can quickly identify suitable settings with \ours and then apply these configurations to the hyperparameters of other GNNs/LLMs.

\end{document}